\pdfoutput=1
\documentclass[11pt]{article}

\usepackage[preprint]{acl}

\usepackage{times}
\usepackage{latexsym}
\usepackage[T1]{fontenc}
\usepackage[utf8]{inputenc}
\usepackage{microtype}
\usepackage{inconsolata}
\usepackage{graphicx}
\usepackage{amsmath}
\usepackage{booktabs}
\usepackage{multirow}
\usepackage{xcolor}

\newcommand{\repourl}{\url{https://github.com/SummitAnthony/BanglaBBQ-benchmark}}

\newcommand{\samb}{s_{\text{AMB}}}
\newcommand{\sdis}{s_{\text{DIS}}}
\newcommand{\diffamb}{\text{diff}_{\text{AMB}}}
\newcommand{\diffdis}{\text{diff}_{\text{DIS}}}
\newcommand{\accamb}{\text{acc}_{\text{amb}}}
\newcommand{\accdis}{\text{acc}_{\text{dis}}}

\title{BanglaBBQ: A Bias Benchmark for Question Answering in the\\Bangladeshi Context}

\author{Sumit Anthony Gomes \and Yasin Rahman \and Taeeba Tasnia \and Faiza Nooren Suhita \\
  BRAC University, Dhaka, Bangladesh \\
  {\small\texttt{sumit.anthony.gomes@g.bracu.ac.bd}\quad\texttt{yasin.rahman@g.bracu.ac.bd}} \\
  {\small\texttt{taeeba.tasnia@g.bracu.ac.bd}\quad\texttt{faiza.nooren.suhita@g.bracu.ac.bd}}}

\begin{document}
\maketitle

\begin{abstract}
This paper presents BanglaBBQ, a question-answering bias benchmark designed to capture stereotypes grounded in the Bangladeshi context, in both Bangla and English. Following KoBBQ and PakBBQ, BanglaBBQ adapts the original Bias Benchmark for Question Answering (BBQ) by classifying its templates into three groups: directly translated (DT), modified to align target groups or contexts with Bangladesh (TM), and removed when deemed unsuitable for the Bangladeshi context (SR). This process retains 140 templates across six categories, comprising 76 DT and 64 TM templates. Five newly authored (NA) categories, based on stereotypes common in Bangladeshi society, are then added: marital status, age and political perception, regional origin, educational background, and mental-health status. Native Bangladeshi annotators judged all 466 newly authored templates, and 372 were retained because a strict majority of English annotators identified the intended target and rated the attitude as common enough. Each newly authored template is linked to a documented source and a verbatim quote from it. Each evaluated template also has a counter-stereotypical version of its disambiguated context. The dataset was evaluated using two open-weight models (Qwen3-4B and Qwen3-8B) and two closed models (GPT-4.1-mini and Gemini 2.5 Flash). All four lean towards the stereotyped person when the context is ambiguous, in both languages, and the effect of language differs by model, so English results do not predict behaviour in Bangla. On the adapted BBQ templates, both Qwen3 models are also less accurate when the context contradicts a stereotype than when it confirms it. On the newly authored templates, the closed models follow counter-stereotypical evidence almost perfectly, and all four models are more biased on the templates that annotators validated than on those they rejected.
\end{abstract}

\section{Introduction}
\label{sec:intro}

Large language models (LLMs) inherit and may amplify the social stereotypes in their training data \citep{gallegos2024bias}. The Bias Benchmark for Question Answering (BBQ) \citep{parrish2022bbq} measures this through multiple-choice questions over ambiguous and disambiguated contexts, but its stereotypes are those of the United States. Culturally adapted variants such as KoBBQ \citep{jin2024kobbq}, CBBQ \citep{huang2024cbbq}, and PakBBQ \citep{hashmat2025pakbbq} show that bias evaluation in another society needs its own stereotypes, groups, and human validation rather than translation alone.

Bangla is one of the world's most widely spoken languages, yet resources for evaluating social bias in it are scarce. Bangla benchmarks such as BLUB \citep{bhattacharjee2022banglabert} target general language understanding, and BharatBBQ \citep{tomar2025bharatbbq} includes Bengali but draws its stereotypes from India. BanStereoSet \citep{kamruzzaman2024banstereoset} localises sentence-level stereotype probes to Bangla.

BanglaBBQ addresses this gap.\footnote{Data and code: \repourl} It keeps BBQ's structure and combines two kinds of templates: newly authored templates addressing the most common stereotypes in Bangladesh, and BBQ's own templates adapted to Bangladesh with the scheme of KoBBQ and PakBBQ. Our contributions are:
\begin{itemize}
  \item 466 newly authored templates in five categories, each linked to a documented source and written with a counter-stereotypical version (usable for 454); 372 retained after judgments by native Bangladeshi annotators (\S\ref{sec:method}--\ref{sec:categories});
  \item an adaptation of BBQ to Bangladesh: 140 templates from six categories, directly translated or target-modified, giving 15,328 instances (\S\ref{sec:adapt});
  \item parallel English and Bangla data for both parts, and an evaluation of two open-weight and two closed LLMs showing that all four rely on stereotypes when the context is ambiguous, that the effect of language differs by model, and that models are more biased on the templates annotators validated (\S\ref{sec:experiments}--\ref{sec:discussion}).
\end{itemize}

An earlier version of BanglaBBQ was introduced in an undergraduate thesis \citep{tasnia2026banglabbq}, which proposed five Bangladesh-specific bias categories and the SafeLLM inference-time mitigation pipeline. The benchmark presented here was rebuilt from scratch and extends that work substantially: every newly authored template is grounded in a verified source quote (144 sources), validated by native annotators, and paired with a counter-stereotypical version; the adapted BBQ templates were re-sorted and re-localized; and all experiments are new.

\section{Related Work}
\label{sec:related}

\paragraph{Bias benchmarks and BBQ.} Stereotype benchmarks such as StereoSet \citep{nadeem2021stereoset}, CrowS-Pairs \citep{nangia2020crows}, and BOLD \citep{dhamala2021bold} measure bias through sentence likelihoods or open-ended generation \citep{gallegos2024bias}. BBQ \citep{parrish2022bbq} instead asks multiple-choice questions about two people. In an ambiguous context, the correct answer is \textsc{unknown}, while in a disambiguated context added evidence points to a correct answer. BBQ includes both stereotype-consistent and stereotype-inconsistent versions. Its 325 templates cover nine U.S.-centric categories and two intersections of them, each grounded in a documented harm, and expand to over 58,000 examples. \citet{parrish2022bbq} caution that BBQ is designed only for U.S. English-speaking contexts and should not be used as evidence that a model would look unbiased in another culture.

\paragraph{Culturally adapted BBQ benchmarks.} KoBBQ \citep{jin2024kobbq} introduced a framework for adaptation: each BBQ template is transferred, target-modified, or removed, and new Korea-specific templates are added, with targets verified by a survey. It also finds that a machine-translated BBQ measures bias differently from the adapted one. PakBBQ \citep{hashmat2025pakbbq} applies the same scheme to Pakistan (directly translated, target-modified, newly added, and simply removed templates), validates the new templates with native annotators, and translates the data into Urdu. CBBQ \citep{huang2024cbbq} builds a Chinese benchmark with human-AI collaboration, BharatBBQ \citep{tomar2025bharatbbq} covers eight languages, including Bengali and English, and MBBQ \citep{neplenbroek2024mbbq} compares stereotypes shared across four languages to separate language from culture.

\paragraph{Research gap.} KoBBQ, CBBQ, and PakBBQ show that a new culture needs locally specific bias dimensions validated by native speakers. Related dimensions such as region, education, and political orientation appear in KoBBQ and CBBQ, but the stereotypes and groups they encode are those of other societies. Although BharatBBQ provides a BBQ-style benchmark in Bengali, it focuses on India rather than Bangladesh. BanglaBBQ addresses this gap with a benchmark grounded in Bangladeshi society, comprising both newly authored and adapted templates.

\section{Methodology}
\label{sec:method}

This section and the next describe the newly authored templates of BanglaBBQ; the adapted BBQ templates are described in \S\ref{sec:adapt}. We construct the newly authored part as a source-grounded benchmark rather than a translation of an existing one. Our pipeline has six stages: (i)~selecting Bangladesh-specific bias categories and target groups; (ii)~drafting templates from documented sources, each with a verbatim quote; (iii)~verifying every source and quote; (iv)~expanding each template into the BBQ/PakBBQ instance format; (v)~translating into Bangla; and (vi)~human annotation in both languages. The design follows the ambiguous/disambiguated QA formulation of BBQ \citep{parrish2022bbq} and the cultural adaptation of KoBBQ, PakBBQ, and CBBQ \citep{jin2024kobbq,hashmat2025pakbbq,huang2024cbbq}.

\subsection{Category and Target Selection}
\label{sec:selection}

Prior adaptations classify each original BBQ template as transferred, target-modified, or removed, and add newly created templates \citep{jin2024kobbq}. Several common stereotypes in Bangladeshi society are not captured in BBQ, so we author new categories for them (newly authored, NA), alongside the adapted BBQ templates (\S\ref{sec:adapt}). Age and political perception builds on BBQ's age dimension, but its templates encode stereotypes about the political attitudes of age groups in Bangladesh. We select five categories that are recurrent sites of discrimination in Bangladeshi society and are documented in journalistic and scholarly sources: marital status, age and political perception, regional origin, educational background, and mental-health status.

Within each category there is a target set (the stereotyped groups) and, for each item, a contrasting non-target. Targets are chosen to reflect the groups actually subject to prejudice in the sources rather than abstract opposites (e.g.\ divorced women and widows for marital status, or indigenous and Bihari communities for regional origin). Each category's targets are described in \S\ref{sec:cat-desc}. Each target maps to an \texttt{X}/\texttt{nonX} semantic label following BBQ's convention. Where an attested group had no existing label (e.g.\ deserted women), we coined one consistent with that convention (\texttt{deserted}/\texttt{nonDeserted}).

\subsection{Template Design}
\label{sec:design}

The unit of construction is a template. Following BBQ \citep{parrish2022bbq}, each template contains an ambiguous context that introduces the target and non-target without enough information to answer, a disambiguating clause that settles who the answer is, and a negative and a non-negative question over three answer options (the two people and \textsc{unknown}). To control for lexical confounds, the target and non-target are the only substantive difference between the two answer strings; answer positions are balanced at evaluation (\S\ref{sec:experiments}).

We first studied BBQ, KoBBQ, and PakBBQ. We then identified stereotypes current in Bangladesh by reading newspapers and scholarly articles, using LLM assistants to survey what has been documented, and consulting anthropology professors. For each category we fixed the target groups and the areas of life to cover, and built the templates with the help of LLM assistants, each from a passage of a public document about Bangladesh; we reviewed and approved every template. The drafting setup, including the instructions given to the assistants, is documented in the repository. The age and political perception templates were written by one author and reviewed by another. Every template carries a source URL and a \texttt{source\_quote}, a passage copied verbatim from that page. Each quote was verified against its source page and checked for whether it supports the encoded stereotype; templates that failed were re-grounded on another verified source or dropped. For age and political perception, the 100 templates share 11 sources that document the political context (young people's participation, elections, protest, corruption, press freedom, and media use) rather than each age stereotype, so in this category the stereotypes rest mainly on the annotators' judgments. Every template was then judged by native annotators (\S\ref{sec:annotation}).

\subsection{Template Expansion}
\label{sec:expansion}

Each template is expanded deterministically into eight test instances, following the block structure of BBQ and PakBBQ \citep{parrish2022bbq,hashmat2025pakbbq}: 2 context conditions (ambiguous, disambiguated) $\times$ 2 question polarities (negative, non-negative) $\times$ 2 answer orderings (target, \textsc{unknown}, non-target; and non-target, target, \textsc{unknown}). The compiled records use the exact BBQ/PakBBQ JSON schema, so they can be read by existing BBQ tooling: each instance stores an \texttt{answer\_info} map with an \texttt{X}/\texttt{nonX} label per option (the \textsc{unknown} option is always tagged \texttt{unknown}), an \texttt{additional\_metadata} block (\texttt{subcategory}, \texttt{stereotyped\_groups}, \texttt{version}, \texttt{source}), the context and question, three answer strings, and the gold label. Our scorer identifies biased answers from these tags rather than by string-matching \texttt{stereotyped\_groups} (\S\ref{sec:experiments}).

By construction the correct answer is \textsc{unknown} on all four ambiguous instances, the target on the negative disambiguated instances, and the non-target on the non-negative disambiguated instances. We enforce three balance invariants programmatically on every compiled category file: a gold-label ratio of exactly $2:3:3$ across the three answer positions, a 50/50 split of ambiguous vs.\ disambiguated instances, and a 50/50 split of negative vs.\ non-negative questions, with each template contributing exactly eight version-A instances with unique identifiers; retained templates keep their original identifiers after validation.

\paragraph{Counter-stereotypical version.} For each template, a second disambiguating clause was written in which the non-target is the answer to the negative question and the target the answer to the non-negative one (version B), as in BBQ \citep{parrish2022bbq}. It yields four further disambiguated instances per template; the ambiguous instances are shared with version A. Version-B clauses were written and checked by the authors and were not annotated. A content review of all version-B items, completed on 11 October 2026 (the last day of the version-B model runs), excluded twelve templates whose questions presuppose something that fits only the stereotyped person or whose counter-stereotypical answer is strained (\S\ref{sec:two-version}).

\subsection{Human Annotation}
\label{sec:annotation}

To confirm that each documented stereotype is recognisable in Bangladesh and points to the intended group, templates were judged by native annotators, using a protocol adapted from PakBBQ \citep{hashmat2025pakbbq}.

\paragraph{Annotators and protocol.} The annotators were native Bangladeshi undergraduate and graduate students from several universities, recruited through a job posting, screened by CV and interview, and paid BDT~1,000 (about US\$8) per hour; none is an author. Each gave written informed consent after a briefing on the study and its harmful content. The English judgments come from three annotators, the same three for all five categories; the Bangla judgments come from a separate group of annotators, three per category, so the English and Bangla judgments of a category come from different annotators.

Annotation was done in Label Studio \citep{tkachenko2020label}, one screen per template. Each screen showed the template's ambiguous context, its negative question, and the two groups. Annotators chose the group the question points to (A, B, or can't tell), judged as most people in Bangladeshi society would perceive it, and rated how common the attitude is in Bangladesh (relevance) on a 1--3 scale. They never saw the disambiguating clause, the source, or the intended answer. Each template received three judgments per language (two for one Bangla template).

\paragraph{Agreement.} After mapping each choice to target, non-target, or can't tell, we measure agreement with Gwet's AC1 \citep{gwet2008computing}, with 95\% bootstrap confidence intervals over templates, and report Fleiss' $\kappa$ \citep{fleiss1971measuring} for comparison (Table~\ref{tab:annotation}). Agreement is moderate to substantial in English (AC1 0.563--0.798) and substantial to almost perfect in Bangla (0.617--0.880), using the bands of \citet{landis1977measurement}. Bangla relevance ratings are nearly uniform (89\% are 3, against 56\% in English). This may reflect broad agreement among the Bangla annotators that these attitudes are common, but it also means that the Bangla ratings discriminate little between templates, so the higher Bangla AC1 should be read with this in mind. Fleiss' $\kappa$ is much lower in several categories, most clearly for Mental-health status ($\kappa = 0.021$ English, 0.028 Bangla), because nearly all labels select the intended target (90\% and 94\%), which pushes $\kappa$'s chance correction towards the observed agreement (the kappa paradox; \citealp{feinstein1990high}). AC1 is robust to this, so we use it as the primary measure.

\paragraph{Retention.} A template is retained if a strict majority of its English raters choose the intended target and its mean relevance is at least 1.5. The English judgments decide because the templates and all their sources are originally in English, and a single decision keeps the English and Bangla sets identical. Of the 466 templates, 372 (79.8\%) are retained; the relevance criterion removed two of them. The 360 retained templates that also have a counter-stereotypical version (\S\ref{sec:two-version}) form the evaluation set.

\begin{table*}[t]
\centering
\small
\resizebox{\textwidth}{!}{%
\begin{tabular}{lrcccccccc}
\toprule
 & & \multicolumn{2}{c}{AC1 [95\% CI]} & \multicolumn{2}{c}{Fleiss' $\kappa$} & \multicolumn{2}{c}{Kept} & \\
\cmidrule(lr){3-4}\cmidrule(lr){5-6}\cmidrule(lr){7-8}
Category & $n$ & EN & BN & EN & BN & EN & BN & Retained \\
\midrule
Marital status             & 100 & 0.654 [0.557, 0.748] & 0.821 [0.753, 0.888] & 0.351 & 0.360 & 80 & 89 & 80 \\
Age \& political perception & 100 & 0.563 [0.463, 0.662] & 0.617 [0.523, 0.711] & 0.446 & 0.438 & 62 & 69 & 62 \\
Regional origin            & 100 & 0.658 [0.564, 0.744] & 0.678 [0.572, 0.772] & 0.224 & 0.221 & 85 & 83 & 85 \\
Educational background     &  95 & 0.667 [0.579, 0.752] & 0.716 [0.614, 0.803] & 0.286 & 0.177 & 77 & 85 & 77 \\
Mental-health status       &  71 & 0.798 [0.702, 0.875] & 0.880 [0.805, 0.942] & 0.021 & 0.028 & 68 & 70 & 68 \\
\midrule
Total                      & 466 & & & & & 372 & 396 & 372 \\
\bottomrule
\end{tabular}}
\caption{Human annotation of BanglaBBQ templates in English (EN) and Bangla (BN). $n$: templates annotated. AC1: Gwet's AC1 on target / non-target / can't tell. Kept: templates on which a strict majority chose the intended target and mean relevance was at least 1.5. Retained: kept in English.}
\label{tab:annotation}
\end{table*}

\subsection{Bangla Translation}
\label{sec:translation}

BanglaBBQ is released as parallel English and Bangla datasets aligned by instance identifier. Because a machine-translated BBQ can measure bias differently from a culturally adapted one \citep{jin2024kobbq}, a native Bangla speaker guided the translation. We translate the template fields and compile both languages identically, keeping the \texttt{answer\_info} group labels in English so that scoring is language-agnostic. The Bangla was produced by LLM-assisted translation following terminology and register choices set by a native-Bangla-speaking author, and several native Bangla speakers then judged its quality and found it acceptable.

\section{The Newly Authored Categories}
\label{sec:categories}

Table~\ref{tab:na-counts} gives the number of templates authored, retained, and evaluated in each newly authored category and the resulting test instances.

\begin{table}[t]
\centering
\small
\resizebox{\columnwidth}{!}{%
\begin{tabular}{lrrrr}
\toprule
Category & Auth. & Ret. & Eval. & Inst. \\
\midrule
Marital status              & 100 & 80 & 75 & 900 \\
Age \& political perception & 100 & 62 & 62 & 744 \\
Regional origin             & 100 & 85 & 80 & 960 \\
Educational background      &  95 & 77 & 76 & 912 \\
Mental-health status        &  71 & 68 & 67 & 804 \\
\midrule
Total                       & 466 & 372 & 360 & 4,320 \\
\bottomrule
\end{tabular}}
\caption{Newly authored templates by category: authored (Auth.), retained after human validation (Ret., \S\ref{sec:annotation}), evaluated (Eval.: retained and with a usable counter-stereotypical version, \S\ref{sec:two-version}), and test instances per language (Inst.) for the evaluated templates (eight in version A and four in version B per template). Five further Educational-background templates were dropped before annotation because their quotes did not support the encoded stereotype.}
\label{tab:na-counts}
\end{table}

\subsection{Categories and Stereotypes}
\label{sec:cat-desc}

Each category targets stereotypes documented in Bangladeshi journalistic and scholarly sources (for age and political perception, see the caveat in \S\ref{sec:design}).

\paragraph{Marital status.} Divorced women, widows, never-married women and men, married women (the ``maternal wall'' at work), women in polygamous marriages, and deserted women, in housing, hiring, inheritance, remarriage, and social judgment.

\paragraph{Age and political perception.} Young adults (18--25), stereotyped as distrustful and disengaged, and middle-aged (41--60) and older adults (60+), stereotyped as deferential to authority, across political trust, participation, protest, and media trust.

\paragraph{Regional origin.} Indigenous/Adivasi people, Biharis, tea-garden workers, Rohingya, char-dwellers, and residents of specific districts (e.g.\ Noakhali, Rangpur, Barishal), in housing, employment, marriage, and blame attribution.

\paragraph{Educational background.} Madrasa-stream graduates (employability, radicalisation suspicion), English-medium students (elitism), and Bangla-medium, low-education, and vocational groups, in hiring, admission, workplace competence, and marriage proposals.

\paragraph{Mental-health status.} People described as depressed, anxious, schizophrenic, or ``mentally ill'', stereotyped as weak-minded, attention-seeking, or dangerous in family, work, healthcare, and marriage.

\subsection{Source Provenance and Diversity}
\label{sec:provenance}

The 466 templates cite 144 distinct URLs on 35 domains. The 100 marital-status templates draw on 57 sources across 19 domains; the other categories draw on 11--32 sources each. Sources include national newspapers (The Daily Star, The Business Standard, Dhaka Tribune, Prothom Alo, New Age), peer-reviewed venues (Springer, Cambridge, PMC/NCBI), rights bodies (Human Rights Watch, Minority Rights Group, UN Women), and policy and development organisations (World Bank, Freedom House, Centre for Policy Dialogue, Transparency International Bangladesh).

\subsection{Release and Intended Use}
\label{sec:release}

We release all 466 templates in the BBQ/PakBBQ JSON format, in parallel English and Bangla files, with each template's source URL, quote, and validation outcome (the 94 rejected templates are marked and used only in \S\ref{sec:valid-rej}), together with the adapted BBQ templates (\S\ref{sec:adapt}), the model outputs, and the code, at \repourl. Data are released under CC BY 4.0, except the verbatim \texttt{source\_quote} excerpts, which remain under their publishers' copyright, and the closed-model outputs, which are released for research use under the providers' terms; code is released under MIT (BBQ-derived scripts: CC BY 4.0). The intended use is set out in the Ethics Statement.

\section{Adapting BBQ to Bangladesh}
\label{sec:adapt}

The second part of BanglaBBQ adapts BBQ's own templates, following KoBBQ and PakBBQ \citep{jin2024kobbq,hashmat2025pakbbq}. The authors sorted each of the 343 templates in BBQ's released template files, which include its two intersectional categories, into three types. A template is directly translated (DT) if its scenario and stereotype both hold in Bangladesh, so that only the language changes; removing U.S.-only entries from a template's word lists (e.g.\ ``jock'') keeps it DT. It is target-modified (TM) if the stereotype holds but an American detail had to be replaced, and removed (SR) if the stereotype does not hold or means something different in Bangladesh. Table~\ref{tab:adapt} gives the result.

\begin{table}[t]
\centering
\small
\begin{tabular}{lrrrr}
\toprule
Category & DT & TM & SR & Instances \\
\midrule
Age                   & 14 &  7 &  4 & 2,488 \\
Disability status     & 20 &  5 &  0 & 1,432 \\
Gender identity       & 13 & 30 &  7 & 4,808 \\
Physical appearance   & 15 &  7 &  3 & 1,296 \\
Religion              &  4 &  1 & 20 &   160 \\
Socio-economic status & 10 & 14 &  1 & 5,144 \\
\midrule
Total                 & 76 & 64 & 35 & 15,328 \\
\bottomrule
\end{tabular}
\caption{Adapted BBQ templates by category. DT: directly translated; TM: target-modified; SR: removed. Instances per language (5,928 DT and 9,400 TM). Five further BBQ categories (168 templates) were removed.}
\label{tab:adapt}
\end{table}

\paragraph{Removed templates (SR).} Five categories were removed entirely. Race/ethnicity and its two intersections with socio-economic status and gender are based on U.S.-specific racial classifications that do not directly apply to Bangladesh. Similarly, the Nationality category focuses on targets and stereotypes related to U.S.\ immigration. The sexual-orientation templates are framed around U.S.\ identities and would need separate, community-led validation, which we leave to future work. Within the six remaining categories, most religion templates were removed because they targeted groups rarely found in Bangladesh, such as Jewish and Mormon communities, or reflected Western perceptions of Muslims. Transgender-related templates in the Gender Identity category were also excluded because the U.S.\ distinction between cisgender and transgender identities does not fully reflect the distinct cultural identity of \emph{hijra} in Bangladesh. Additionally, individual templates were removed when their stereotypes did not align with the Bangladeshi context, such as associating beards with poverty, as beards can serve as religious markers in Bangladesh.

\paragraph{Target-modified templates.} TM edits replace American details with Bangladeshi equivalents while keeping the stereotype: for example, Walmart becomes the supermarket chain Shwapno, 911 becomes 999, Jeopardy becomes a TV quiz show, ``on welfare'' becomes ``living on government relief'', and cocaine becomes yaba. BBQ's U.S.\ first names are replaced by 60 common Bangladeshi names (Muslim and Hindu, women and men), with both people in a pair sharing a religion, and U.S.-only entries are dropped from shared lists such as occupations and religions.

\paragraph{Generation.} We regenerate the instances with BBQ's own generation script, so the adapted templates keep BBQ's structure: blocks of four instances, both orders of the two people, and both a stereotype-consistent and a stereotype-inconsistent version of every disambiguated context, split 50/50 within each template. As in PakBBQ, the unknown option is always ``Unknown''. The files follow PakBBQ's layout, one per category with DT instances before TM instances.

\paragraph{Translation and validation.} An LLM translated the adapted templates and slot fillers into Bangla with the conventions of \S\ref{sec:translation}, and native Bangla speakers judged the translation acceptable; only the context, question, and answer strings are translated, and the Bangla unknown option is ``not enough information'' (\S\ref{sec:experiments}). As in PakBBQ, which validated only its newly added templates, the adapted templates were not annotated.

\section{Experiments}
\label{sec:experiments}

\subsection{Experimental Setup}
\label{sec:setup}

\paragraph{Models.} We evaluate two open-weight and two closed models: Qwen3-4B and Qwen3-8B \citep{qwen2025qwen3}, GPT-4.1-mini \citep{openai2025gpt41}, and Gemini 2.5 Flash \citep{comanici2025gemini}. Unless noted (Table~\ref{tab:wording}), the Qwen3 models were run with Hugging Face Transformers on Colab and Kaggle GPUs with greedy decoding and their reasoning mode disabled; the closed models were queried through the OpenRouter API (\texttt{openai/gpt-4.1-mini}, \texttt{google/gemini-2.5-flash}) at temperature 0, with Gemini's thinking disabled. All runs took place between 29 September and 11 October 2026.

\paragraph{Prompting.} We evaluate zero-shot, without a system prompt or persona; the prompt gives the context, the question, and options A--C and asks for a single letter, and the Bangla prompt is a translation of the English one (Appendix~\ref{sec:prompts}). Each item is presented in all three cyclic orders of its options \citep{jin2024kobbq} and answered by majority vote \citep{hashmat2025pakbbq}; items with no majority or an answer outside A--C count as incorrect. The two answer orderings of a template (\S\ref{sec:expansion}) are cyclic rotations of each other, so under this protocol they yield the same three presentations. The Bangla unknown option is ``not enough information'', one of BBQ's phrasings, because a literal translation of ``unknown'' was almost never chosen in a pilot.

\paragraph{Metrics.} We report ambiguous and disambiguated accuracy and the BBQ ambiguous bias score \citep{parrish2022bbq}
\begin{equation*}
  \samb = (1 - \accamb)\left(2\,\frac{n_{\text{biased}}}{n_{\neg\text{unk}}} - 1\right),
\end{equation*}
where $n_{\neg\text{unk}}$ counts valid non-unknown answers (items without a valid answer count as incorrect in $\accamb$ but are not in $n_{\neg\text{unk}}$) and an answer is biased if it names the target for a negative question or the non-target for a non-negative one. We also report KoBBQ's $\diffamb = (n_{\text{biased}} - n_{\text{counter}})/N_{\text{amb}}$, where $n_{\text{counter}}$ counts counter-stereotypical answers and $N_{\text{amb}}$ all ambiguous items, and $\diffdis$ (disambiguated accuracy on stereotype-consistent minus counter-stereotypical contexts; \citealp{jin2024kobbq}), BBQ's $\sdis = 2\,n_{\text{biased}}/n_{\neg\text{unk}} - 1$ on disambiguated items, the share of all items (both versions) without a valid answer (OOS), and the share answered identically in all three orders (consistency). Positive bias scores are stereotype-aligned. Intervals are 95\% bootstrap intervals over templates (1,000 resamples).

\subsection{Two-Version Evaluation Design}
\label{sec:two-version}

In version A of the newly authored templates, every disambiguating clause confirms the stereotype, so $\sdis$ is close to $+1$ for any model that reads the context and measures comprehension rather than bias. Version B adds, for each template, a counter-stereotypical disambiguating clause in which the non-target is the answer to the negative question and the target to the non-negative one, as in BBQ \citep{parrish2022bbq}. Ambiguous instances are shared by both versions, so ambiguous-context scores do not depend on the version; with both versions, $\sdis$ and $\diffdis$ measure whether a model still prefers the stereotyped answer when the evidence says otherwise.

Twelve templates have no usable version B, because their questions presuppose something that fits only the stereotyped person or the counter-stereotypical answer is strained. The evaluation set therefore consists of the 360 templates that were validated by annotators (\S\ref{sec:annotation}) and have a usable version B, with 2,880 version-A and 1,440 version-B instances per language.

\begin{table*}[t]
\centering
\small
\setlength{\tabcolsep}{4.5pt}
\resizebox{\textwidth}{!}{%
\begin{tabular}{llccccccccc}
\toprule
Model & & $\accamb$ [95\% CI] & $\accdis$ & $\samb$ [95\% CI] & $\diffamb$ & $\sdis$ & $\diffdis$ [95\% CI] & OOS & Consist. \\
\midrule
\multirow{2}{*}{Qwen3-4B}         & EN & 0.169 [0.14, 0.20] & 0.992 & $+$0.425 [0.36, 0.49] & $+$0.407 & $+$0.012 & $+$0.012 [0.00, 0.02] & 1.2\% & 0.862 \\
                                  & BN & 0.069 [0.05, 0.09] & 0.963 & $+$0.336 [0.27, 0.40] & $+$0.328 & $+$0.042 & $+$0.042 [0.03, 0.06] & 0.7\% & 0.825 \\
\midrule
\multirow{2}{*}{Qwen3-8B}         & EN & 0.461 [0.42, 0.51] & 0.999 & $+$0.293 [0.24, 0.35] & $+$0.278 & $+$0.001 & $+$0.001 [0.00, 0.00] & 0.9\% & 0.866 \\
                                  & BN & 0.554 [0.51, 0.60] & 0.988 & $+$0.210 [0.16, 0.26] & $+$0.176 & $+$0.013 & $+$0.017 [0.01, 0.03] & 2.6\% & 0.839 \\
\midrule
\multirow{2}{*}{GPT-4.1-mini}     & EN & 0.372 [0.33, 0.41] & 0.999 & $+$0.475 [0.42, 0.52] & $+$0.469 & $+$0.002 & $+$0.003 [0.00, 0.01] & 0.3\% & 0.910 \\
                                  & BN & 0.384 [0.35, 0.43] & 0.993 & $+$0.399 [0.34, 0.45] & $+$0.396 & $+$0.002 & $+$0.002 [$-$0.00, 0.01] & 0.2\% & 0.897 \\
\midrule
\multirow{2}{*}{Gemini 2.5 Flash} & EN & 0.798 [0.76, 0.83] & 0.999 & $+$0.146 [0.11, 0.18] & $+$0.142 & 0.000 & $-$0.001 [$-$0.00, 0.00] & 0.2\% & 0.940 \\
                                  & BN & 0.690 [0.65, 0.73] & 0.998 & $+$0.218 [0.17, 0.26] & $+$0.209 & $+$0.001 & $+$0.001 [0.00, 0.00] & 0.4\% & 0.916 \\
\bottomrule
\end{tabular}}
\caption{Results on the newly authored templates (360 templates, versions A and B). Positive bias scores are stereotype-aligned. $\sdis$ and $\diffdis$ are computed over versions A and B together. OOS: share of all items with no valid majority answer. Consist.: share of items answered identically in all three option orders.}
\label{tab:main}
\end{table*}

\begin{table}[t]
\centering
\small
\setlength{\tabcolsep}{4pt}
\begin{tabular}{llccccc}
\toprule
Model & & Mar. & Age-Pol. & Reg. & Edu. & MH \\
\midrule
\multirow{2}{*}{Qwen3-4B}      & EN & .413 & .347 & .200 & \textbf{.763} & .299 \\
                               & BN & .347 & .315 & .000 & .447 & \textbf{.575} \\
\multirow{2}{*}{Qwen3-8B}      & EN & .187 & .234 & .131 & \textbf{.493} & .351 \\
                               & BN & .113 & .089 & .025 & \textbf{.342} & .321 \\
\multirow{2}{*}{GPT-4.1-mini}  & EN & .470 & .415 & .494 & \textbf{.612} & .325 \\
                               & BN & .453 & .351 & .353 & \textbf{.520} & .284 \\
\multirow{2}{*}{Gemini 2.5 F.} & EN & .193 & .016 & \textbf{.212} & .099 & .164 \\
                               & BN & \textbf{.287} & .024 & .256 & .197 & .250 \\
\bottomrule
\end{tabular}
\caption{Ambiguous-context bias $\diffamb$ by category (highest per row in bold). Mar.: marital status (75 templates); Age-Pol.: age and political perception (62); Reg.: regional origin (80); Edu.: educational background (76); MH: mental-health status (67).}
\label{tab:category}
\end{table}

\subsection{Main Results: Ambiguous-Context Bias}
\label{sec:main-results}

Table~\ref{tab:main} gives the overall results. Every model has a positive ambiguous-context bias score in both languages ($\samb$ $+0.146$ to $+0.475$): when a model does not answer unknown, it tends to name the stereotyped person.
\begin{itemize}
  \item Gemini 2.5 Flash answers \textsc{unknown} most often ($\accamb$ 0.798 English, 0.690 Bangla) and is the least biased model in English; in Bangla its bias is level with Qwen3-8B's ($\samb$ $+0.218$ vs.\ $+0.210$). Its bias is still clearly positive ($\diffamb$ $+0.142$, $+0.209$).
  \item GPT-4.1-mini is the most biased model ($\samb$ $+0.475$ English, $+0.399$ Bangla).
  \item Within the Qwen3 family, the larger model is less biased ($+0.293$ vs.\ $+0.425$ in English, $+0.210$ vs.\ $+0.336$ in Bangla).
  \item All models answer disambiguated items almost perfectly ($\accdis$ 0.963--0.999).
\end{itemize}

\paragraph{Categories.} Table~\ref{tab:category} breaks the ambiguous-context bias down by category. Educational background has the highest bias for both Qwen3 models and for GPT-4.1-mini in both languages, except for Qwen3-4B in Bangla, where mental-health status is highest. Educational-background bias reaches $+0.763$ for Qwen3-4B and $+0.612$ for GPT-4.1-mini in English. Gemini 2.5 Flash is most biased on Regional origin in English and Marital status in Bangla. Age \& political perception has the lowest bias for Gemini 2.5 Flash ($+0.016$ English, $+0.024$ Bangla), and both Qwen3 models show almost no regional bias in Bangla (0.000 and $+0.025$) despite clear regional bias in English.

\paragraph{English and Bangla.} The effect of language differs by model. Bias is lower in Bangla than in English for Qwen3-4B, Qwen3-8B, and GPT-4.1-mini ($\samb$ lower by 0.089, 0.082, and 0.076) but higher in Bangla for Gemini 2.5 Flash (by 0.072). These differences hold for prompt wording~1; for GPT-4.1-mini the direction reverses under wording~3 (Table~\ref{tab:wording}). Ambiguous-context accuracy is lower in Bangla for Qwen3-4B and Gemini 2.5 Flash, higher for Qwen3-8B, and about the same for GPT-4.1-mini. English results therefore do not predict how a model behaves in Bangla.

\subsection{Validated and Rejected Templates}
\label{sec:valid-rej}

Human validation also separates templates on which models are biased. Table~\ref{tab:valid-rej} (Appendix~\ref{sec:appendix}) compares the 372 templates that annotators validated with the 94 they rejected. For every model and language, bias is higher and ambiguous-context accuracy lower on the validated templates; for GPT-4.1-mini, $\samb$ is $+0.475$ on validated and $+0.179$ on rejected templates in English. The confidence intervals of the two sets do not overlap in seven of the eight model--language pairs (all but Qwen3-4B in Bangla). The rejected set is small and unevenly spread across categories (38 Age \& political perception, 20 Marital, 18 Educational, 15 Regional, and 3 Mental-health templates), so per-category comparisons are not reliable, and the two sets also differ in their category mix.

\subsection{Prompt-Wording Robustness}
\label{sec:wording}

We reran the validated templates with two further prompt wordings, one of which asks the model to answer ``based only on the context above'' (Table~\ref{tab:wording} in Appendix~\ref{sec:appendix}). Wording changes the size of the bias considerably: GPT-4.1-mini's English $\samb$ ranges from $+0.264$ to $+0.475$, and its Bangla score from $+0.141$ to $+0.402$. The direction does not change, but the ranking can: every score is positive and GPT-4.1-mini is more biased than Gemini 2.5 Flash under every wording (barely so in Bangla with wording 2, $+0.141$ vs.\ $+0.140$), yet in Bangla Qwen3-8B is the least biased of the three models with wording 1 and the most biased with wording 2, although its wording-2 and -3 runs were made mostly through a different inference backend (Table~\ref{tab:wording}). Bias scores should therefore be compared only under the same prompt.

\begin{table}[t]
\centering
\small
\resizebox{\columnwidth}{!}{%
\begin{tabular}{llcccccc}
\toprule
 & & \multicolumn{2}{c}{Accuracy} & & & & \\
\cmidrule(lr){3-4}
Model & & Amb. & Dis. & $\samb$ & $\diffamb$ & $\sdis$ & $\diffdis$ \\
\midrule
\multirow{2}{*}{Qwen3-4B} & EN & .643 & .899 & $+$.224 & $+$.211 & $+$.035 & $+$.044 \\
                          & BN & .323 & .838 & $+$.294 & $+$.275 & $+$.058 & $+$.060 \\
\multirow{2}{*}{Qwen3-8B} & EN & .858 & .904 & $+$.106 & $+$.101 & $+$.026 & $+$.039 \\
                          & BN & .707 & .859 & $+$.178 & $+$.151 & $+$.056 & $+$.063 \\
\bottomrule
\end{tabular}}
\caption{Results on the adapted BBQ templates (140 templates, balanced sample of 3,256 instances per language). $\diffdis$ intervals exclude zero for both models in both languages.}
\label{tab:adapted-results}
\end{table}

\subsection{Disambiguated Contexts: Version A and Version B}
\label{sec:disamb}

Table~\ref{tab:versions} (Appendix~\ref{sec:appendix}) reports disambiguated results for each version. In version A every model reaches an accuracy of at least 0.983, and $\sdis$ is close to $+1$. In version B, where the context names the non-stereotyped person, accuracy stays at 0.979 or higher for all models except Qwen3-4B in Bangla, and $\sdis$ flips to close to $-1$. The two closed models and Qwen3-8B in English therefore follow explicit counter-stereotypical evidence almost perfectly ($\diffdis$ between $-0.001$ and $+0.003$). The smaller model does not fully do so: Qwen3-4B's accuracy falls from 0.983 to 0.942 in Bangla ($\diffdis$ $+0.042$ [0.03, 0.06]), and Qwen3-8B in Bangla also shows a small gap ($+0.017$ [0.01, 0.03]).

\subsection{Adapted BBQ Templates (DT/TM)}
\label{sec:adapted-eval}

We evaluate the two Qwen3 models on the 140 adapted BBQ templates (\S\ref{sec:adapt}), using a balanced sample of 3,256 instances per language with equal numbers of stereotype-consistent and counter-stereotypical blocks per template, drawn after excluding, as in BBQ's analysis, 16 instances that pair two people of the same gender. The closed models were not run on these templates.

Both models again have positive ambiguous-context bias in both languages (Table~\ref{tab:adapted-results}). Disambiguated accuracy is 0.838--0.904, and both models are less accurate when the context contradicts the stereotype than when it confirms it ($\diffdis$ $+0.039$ to $+0.063$). Unlike on the newly authored templates, both models are more biased in Bangla than in English. By category (Table~\ref{tab:adapted-category}, Appendix~\ref{sec:appendix}), Gender identity has the lowest bias for both models; Religion the highest, on only five templates.

\section{Discussion}
\label{sec:discussion}

\paragraph{Bias under ambiguity and under evidence.} On the newly authored templates, the closed models' bias appears almost only when the context is silent: they answer stereotype-consistent and counter-stereotypical disambiguated items equally well, yet name the stereotyped person in ambiguous contexts. Both Qwen3 models in Bangla (Qwen3-4B clearly, Qwen3-8B slightly), and both Qwen3 models on the adapted BBQ templates, are also less accurate when the evidence contradicts the stereotype, so for them the stereotype affects answers that the context settles.

\paragraph{Validation and local stereotypes.} Models are more biased on the templates that annotators recognised as stereotypes than on those they rejected, for every model and language. This suggests that the stereotypes Bangladeshi annotators recognise are also the ones models reproduce, although the two sets also differ in category mix, so the comparison is not controlled. For the same models, bias is also higher on the newly authored templates than on the adapted BBQ templates (Qwen3-8B English $\samb$ $+0.293$ vs.\ $+0.106$); the two sets differ in categories and design, so this comparison is descriptive, but it is consistent with an adapted BBQ alone not covering stereotypes specific to Bangladesh.

\section{Conclusion}
\label{sec:conclusion}

We presented BanglaBBQ, a bias benchmark for Bangla and English that combines (i)~source-grounded, newly authored categories with stereotype-aligned and counter-stereotypical disambiguation and (ii)~BBQ templates adapted to Bangladesh. Across four models, (1)~every model leans towards the stereotyped person in ambiguous contexts in both languages ($\samb$ $+0.146$ to $+0.475$), with bias concentrated in different categories for different models; (2)~on the newly authored templates, the closed models follow counter-stereotypical evidence almost perfectly, while the Qwen3 models do not fully do so in Bangla, Qwen3-4B most clearly; (3)~the effect of language differs by model and template set; (4)~models are more biased on the templates annotators validated; and (5)~prompt wording changes the size of the bias but not its direction. Future work will validate the Bangla templates in a further annotation round, evaluate Bangla-specific models, and run the closed models on the adapted BBQ templates.

\section*{Limitations}

\paragraph{Translated Bangla data.} All templates, newly authored and adapted, were written in English and then translated into Bangla with an LLM. Although native Bangla speakers judged the translations acceptable, and annotators judged the Bangla context and question of every newly authored template (\S\ref{sec:annotation}), the Bangla items are machine translations rather than text written directly in Bangla, and may not read as naturally or carry exactly the same connotations as native-authored items.

\paragraph{Coverage.} BanglaBBQ's newly authored part covers five categories; gender identity, religion, and disability are covered only through adapted BBQ templates, which encode few Bangladesh-specific stereotypes about them (for example, only five Religion templates remain). Every template contrasts one stereotyped group with one reference group along a single dimension, so intersectional biases (e.g.\ age combined with political affiliation, or gender combined with region) are not tested.

\paragraph{Annotator pool.} Retention rests on the English judgments of three university students, the same three for all five categories; the annotators were mostly aged 20--25. They are native speakers, but younger and more educated than the Bangladeshi population as a whole, and their judgments of which stereotypes are common may differ from those of other groups, for example rural or older Bangladeshis.
\paragraph{Retention rule.} We fixed the English-only retention rule after the first model runs; it replaced a rule that pooled the two languages for Age and political perception. Under rules based on the Bangla judgments or on both languages, validated templates remain more biased than rejected ones for every model and language, but the confidence intervals overlap for two or three of the eight model--language pairs. The English annotation that decides retention was completed on 11--12 October 2026, after the version-A model runs (29 September to 10 October), although annotators saw no model outputs; the twelve version-B exclusions (\S\ref{sec:two-version}) were made on 11 October, the last day of the version-B runs.

\paragraph{Unannotated clauses.} Annotators saw only the ambiguous context and the negative question of each template. The disambiguating clauses of versions A and B and the non-negative questions were checked by the authors only.

\paragraph{Changing stereotypes.} Perceptions of stereotypes are subjective and change over time. This is especially true of political attitudes, which can shift quickly, as after Bangladesh's change of government in 2024. Our annotations reflect the views of our annotators in September and October 2026.

\paragraph{Prompt sensitivity.} The size of the bias depends strongly on the prompt wording: GPT-4.1-mini's English $\samb$ ranges from $+0.264$ to $+0.475$ across three wordings (\S\ref{sec:wording}). The direction of the bias holds across wordings, but the ranking of models can change, so scores should only be compared under the same prompt. One Qwen3-8B run failed and was excluded, and Qwen3-4B is reported on wording 1 only because its wording 2 and 3 runs gave many invalid answers (21\% and 62\% of English calls) and its Bangla run was incomplete.

\paragraph{Unknown wording.} The Bangla unknown option was chosen in a pilot that ran Qwen3-4B, one of the evaluated models, on 150 benchmark items (30 of them from a category that was later withdrawn). Of the wordings tested, we chose the one whose Unknown rate was closest to that of English ``Unknown''. English--Bangla differences in ambiguous-context accuracy therefore depend partly on this choice, and Qwen3-4B's Bangla results are not fully independent of it.

\paragraph{Models.} We evaluate four models, two open-weight and two closed, and no Bangla-specific language model: in a pilot, TituLLM-3B never answered with an option letter and TigerLLM-9B did so only in English. Our findings may not generalise to other model families or sizes. The closed models were not evaluated on the adapted BBQ templates.

\paragraph{Scope.} BanglaBBQ measures stereotype reliance in short, templated multiple-choice questions. It does not measure bias in open-ended generation or in real applications. It also covers only standard written Bangla, not regional dialects such as Sylheti or Chittagonian, romanised Bangla, or Bengali as used in India.

\section*{Ethics Statement}

\paragraph{Annotation.} All annotators were paid BDT 1,000 (about US\$8) per hour. They were briefed on the aims of the study and on the harmful content of the items, gave written informed consent, and could pause or stop at any time; their answers are released only as per-template counts. The annotation was not submitted to an institutional ethics board; we treated it as minimal-risk, since adult annotators judged short fictional scenarios and no personal data are released.

\paragraph{Harmful content and intended use.} BanglaBBQ deliberately encodes real, harmful stereotypes about groups in Bangladesh, including marginalised communities such as indigenous people, Biharis, Rohingya refugees, and people with mental illness. It is intended solely for auditing and evaluating bias in language models. It should not be used to train or fine-tune models to reproduce these stereotypes, and a low bias score on BanglaBBQ should not be taken as evidence that a model is unbiased. Taken out of context, the templates could reinforce the stereotypes they document; we release them with this warning and with the source of each template.

\paragraph{Use of AI assistants.} LLM assistants helped us survey documented stereotypes and draft the templates, which we reviewed and approved (\S\ref{sec:design}). An LLM produced the Bangla translation, which native Bangla speakers judged acceptable (\S\ref{sec:translation}), and LLMs were used to help write the code and refine the writing of this paper.

\section*{Acknowledgments}

We thank the authors of BBQ \citep{parrish2022bbq}, whose benchmark, templates, and generation code form the foundation of this work, and the authors of PakBBQ \citep{hashmat2025pakbbq}, whose adaptation of BBQ to Pakistan and annotation protocol we followed closely in building BanglaBBQ. We are also grateful to the KoBBQ team \citep{jin2024kobbq} for the adaptation framework on which both PakBBQ and BanglaBBQ build. Finally, we thank our annotators for their time and careful judgments, and for working through material that encodes harmful stereotypes.

\bibliography{custom}

\appendix

\section{Additional Results}
\label{sec:appendix}

\begin{table}[h]
\centering
\small
\begin{tabular}{llcccc}
\toprule
 & & \multicolumn{2}{c}{$\samb$} & \multicolumn{2}{c}{$\accamb$} \\
\cmidrule(lr){3-4}\cmidrule(lr){5-6}
Model & & Valid. & Rej. & Valid. & Rej. \\
\midrule
\multirow{2}{*}{Qwen3-4B}      & EN & $+$.418 & $+$.218 & .172 & .346 \\
                               & BN & $+$.329 & $+$.197 & .071 & .122 \\
\multirow{2}{*}{Qwen3-8B}      & EN & $+$.290 & $+$.127 & .458 & .777 \\
                               & BN & $+$.211 & $+$.007 & .556 & .761 \\
\multirow{2}{*}{GPT-4.1-mini}  & EN & $+$.475 & $+$.179 & .368 & .593 \\
                               & BN & $+$.402 & $+$.184 & .380 & .598 \\
\multirow{2}{*}{Gemini 2.5 F.} & EN & $+$.148 & $+$.026 & .792 & .918 \\
                               & BN & $+$.219 & $+$.050 & .684 & .894 \\
\bottomrule
\end{tabular}
\caption{Ambiguous-context results on templates that annotators validated (372) and rejected (94), version A.}
\label{tab:valid-rej}
\end{table}

\begin{table}[h]
\centering
\small
\begin{tabular}{llccc}
\toprule
 & & \multicolumn{3}{c}{$\samb$ by wording} \\
\cmidrule(lr){3-5}
Model & & 1 & 2 & 3 \\
\midrule
\multirow{2}{*}{Qwen3-8B}      & EN & $+$.290 & $+$.260 & -- \\
                               & BN & $+$.211 & $+$.200 & $+$.132 \\
\multirow{2}{*}{GPT-4.1-mini}  & EN & $+$.475 & $+$.315 & $+$.264 \\
                               & BN & $+$.402 & $+$.141 & $+$.311 \\
\multirow{2}{*}{Gemini 2.5 F.} & EN & $+$.148 & $+$.094 & $+$.044 \\
                               & BN & $+$.219 & $+$.140 & $+$.106 \\
\bottomrule
\end{tabular}
\caption{Ambiguous-context bias under three prompt wordings (372 validated templates, version A). Wording 1 is used for the main results. The Qwen3-8B English run with wording 3 failed (57\% of items without a valid answer) and is excluded. Qwen3-4B is reported on wording 1 only: its wording 2 and 3 runs gave many invalid answers and were excluded. Qwen3-8B wordings 2 and 3 were run mostly (78\% of the calls scored here) through OpenRouter (\texttt{qwen/qwen3-8b}, temperature 0, no reasoning tokens) and the rest locally; wording 1 was run entirely locally.}
\label{tab:wording}
\end{table}

\begin{table}[h]
\centering
\small
\setlength{\tabcolsep}{4pt}
\resizebox{\columnwidth}{!}{%
\begin{tabular}{llccccc}
\toprule
 & & \multicolumn{2}{c}{Version A} & \multicolumn{2}{c}{Version B} & \\
\cmidrule(lr){3-4}\cmidrule(lr){5-6}
Model & & Acc. & $\sdis$ & Acc. & $\sdis$ & $\diffdis$ \\
\midrule
\multirow{2}{*}{Qwen3-4B}      & EN & .999 & $+$.997 & .986 & $-$.972 & $+$.012 \\
                               & BN & .983 & $+$.967 & .942 & $-$.883 & $+$.042 \\
\multirow{2}{*}{Qwen3-8B}      & EN & 1.00 & $+$1.00 & .999 & $-$.997 & $+$.001 \\
                               & BN & .996 & $+$.992 & .979 & $-$.972 & $+$.017 \\
\multirow{2}{*}{GPT-4.1-mini}  & EN & 1.00 & $+$1.00 & .997 & $-$.997 & $+$.003 \\
                               & BN & .994 & $+$.992 & .992 & $-$.986 & $+$.002 \\
\multirow{2}{*}{Gemini 2.5 F.} & EN & .999 & $+$1.00 & .999 & $-$1.00 & $-$.001 \\
                               & BN & .999 & $+$.997 & .998 & $-$.996 & $+$.001 \\
\bottomrule
\end{tabular}}
\caption{Disambiguated accuracy and $\sdis$ under version A (stereotype-aligned) and version B (counter-stereotypical), 360 templates. $\sdis$ near $+1$ in version A reflects reading comprehension, not bias.}
\label{tab:versions}
\end{table}

\begin{table}[h]
\centering
\small
\resizebox{\columnwidth}{!}{%
\begin{tabular}{lrcccc}
\toprule
 & & \multicolumn{2}{c}{Qwen3-4B} & \multicolumn{2}{c}{Qwen3-8B} \\
\cmidrule(lr){3-4}\cmidrule(lr){5-6}
Category & $n$ & EN & BN & EN & BN \\
\midrule
Age                 & 21 & $+$.211 & $+$.125 & $+$.202 & $+$.098 \\
Disability status   & 25 & $+$.261 & $+$.422 & $+$.072 & $+$.218 \\
Gender identity     & 43 & $+$.055 & $+$.055 & $+$.015 & $+$.026 \\
Physical appearance & 22 & $+$.241 & $+$.241 & $+$.122 & $+$.244 \\
Religion            &  5 & $+$.300 & $+$.800 & $+$.300 & $+$.350 \\
SES                 & 24 & $+$.247 & $+$.443 & $+$.051 & $+$.115 \\
\bottomrule
\end{tabular}}
\caption{Ambiguous-context bias $\diffamb$ on the adapted BBQ templates by category; $n$: templates. Religion has only five templates.}
\label{tab:adapted-category}
\end{table}

\section{Prompts}
\label{sec:prompts}

Figure~\ref{fig:prompts} gives the three English prompt wordings. Each prompt is filled with the item's context, question, and three options, and every item is asked in all three cyclic orders of its options. Wording~1 is used for the main results; wordings~2 and~3 are used only in \S\ref{sec:wording}. The Bangla prompts are translations of these and are given in the released evaluation code.

\begin{figure*}[t]
\centering
\footnotesize
\fbox{\begin{minipage}[t][15.5\baselineskip][t]{0.3\textwidth}\raggedright
\textbf{Wording 1} (main results)\par\smallskip
\ttfamily Read the context and answer the question by choosing one of the options.\\[0.6ex]
Context: \{context\}\\
Question: \{question\}\\
A.~\{a\}\\
B.~\{b\}\\
C.~\{c\}\\[0.6ex]
Answer with only the letter A, B, or C.
\end{minipage}}\hfill
\fbox{\begin{minipage}[t][15.5\baselineskip][t]{0.3\textwidth}\raggedright
\textbf{Wording 2}\par\smallskip
\ttfamily Context: \{context\}\\[0.6ex]
Question: \{question\}\\[0.6ex]
Options:\\
A.~\{a\}\\
B.~\{b\}\\
C.~\{c\}\\[0.6ex]
Based only on the context above, which option is correct? Reply with a single letter (A, B, or C).
\end{minipage}}\hfill
\fbox{\begin{minipage}[t][15.5\baselineskip][t]{0.3\textwidth}\raggedright
\textbf{Wording 3}\par\smallskip
\ttfamily Here is a short passage followed by a multiple-choice question.\\[0.6ex]
Passage: \{context\}\\[0.6ex]
\{question\}\\
(A)~\{a\}\\
(B)~\{b\}\\
(C)~\{c\}\\[0.6ex]
Give your answer as one letter: A, B, or C.
\end{minipage}}
\caption{The three English prompt wordings (\S\ref{sec:setup}, \S\ref{sec:wording}). Line breaks inside a line are due to typesetting.}
\label{fig:prompts}
\end{figure*}

\end{document}